\documentclass[letterpaper, 10 pt, conference]{ieeeconf}  

\IEEEoverridecommandlockouts                              

\usepackage{amsmath} 
\usepackage{amssymb}

\usepackage{ltl} 
\usepackage{multirow}
\usepackage[vlined]{algorithm2e}
\usepackage{mathtools}

\usepackage{color}
\usepackage{subfig}
\usepackage{tikz}
\usepackage{hyperref}
\usetikzlibrary{arrows,fit,shapes,automata}
\usetikzlibrary{positioning,fit,calc,shapes}
\usetikzlibrary{decorations.fractals}
\usetikzlibrary{decorations.markings}
\usepackage{stmaryrd}

\newcommand{\comment}[1]{}

\newtheorem{example}{Example}
\newtheorem{theorem}{Theorem}

\newcommand*{\qed}{\hfill\ensuremath{\blacksquare}}

\newcommand{\init}{\mathsf{init}}
\newcommand{\belief}{\mathsf{belief}}

\newcommand{\vis}{\mathit{vis}}
\newcommand{\Vis}{\mathit{Vis}}
\newcommand{\succs}{\mathit{succ}}
\newcommand{\beliefs}{\mathcal{P}(L_t)}

\newcommand{\states}{S}
\newcommand{\trans}{T}
\newcommand{\part}{\mathcal{Q}}

\newcommand{\post}{\mathit{succ}_t}

\newcommand{\outcome}{\mathit{outcome}}

\newcommand{\bools}{\mathbb{B}}
\newcommand{\true}{\mathit{true}}
\newcommand{\false}{\mathit{false}}
\newcommand{\nats}{\mathbb{N}}

\newcommand{\SP}{\mathcal{SP}}

\newcommand{\locspec}{\mathit{local}}

\title{\LARGE \bf Distributed Synthesis of Surveillance Strategies for Mobile Sensors}

\author{Suda Bharadwaj$^{1}$ and Rayna Dimitrova$^{2}$ and Ufuk Topcu$^{1}$
\thanks{$^{1}$Suda Bharadwaj and Ufuk Topcu are with the University of Texas at Austin}%
\thanks{$^{2}$Rayna Dimitrova is with the University of Leicester, UK.}%
}

\begin{document}

\maketitle
\thispagestyle{empty}
\pagestyle{empty}

\everypar{\looseness=-1}
\begin{abstract}
We study the problem of synthesizing strategies for a mobile sensor network to conduct surveillance in partnership with static alarm triggers. We formulate the problem as a multi-agent reactive synthesis problem with surveillance objectives specified as temporal logic formulas. In order to avoid the state space blow-up arising from a centralized strategy computation, we propose a method to \emph{decentralize} the surveillance strategy synthesis by decomposing the multi-agent game into subgames that can be solved independently. We also decompose the global surveillance specification into local specifications for each sensor, and show that if the sensors satisfy their local surveillance specifications, then the sensor network as a whole will satisfy the global surveillance objective. Thus, our method is able to guarantee global surveillance properties in a mobile sensor network while synthesizing completely decentralized strategies with no need for coordination between the sensors. We also present a case study in which we demonstrate an application of decentralized surveillance strategy synthesis.
\end{abstract}


\section{INTRODUCTION}
The importance of surveillance in our daily life has been constantly growing in the past couple of decades, and with that, also the need for more efficient and sophisticated mechanisms for surveillance. One of the major challenges comes from the need to perform surveillance in large and complex environments, where it is not always feasible or cost effective to have complete surveillance coverage of the entire area at all times. Furthermore, sensors  might not always be able to classify threats, and often require human intervention to assess the threat level. It can thus be necessary to deploy multiple mobile sensors, that work together with conventional static sensors to maintain a sufficient level of knowledge on the location of a potential threat. This is particularly crucial in applications where it is necessary to monitor a potential threat which can move over a large area until it can be accounted for.

In a formal setting, designing a surveillance strategy for a (mobile) sensor network dealing with a potentially adversarial target can be modelled as a two-player game in which one player represents the sensor network and the other player represents the adversary. There are several variants of such games, including pursuit-evasion games~\cite{Chung2011} and graph-searching games~\cite{Kreutzer11}. In such games,	 the problem is formulated as enforcing eventual detection, which is, in essence, a search problem -- once the target is detected, the game ends. These types of games are too restrictive for applications where the goal is not to capture, but instead  to maintain information about the location of the adversary for an unbounded time horizon.

Another class of games used in physical security are \emph{Stackelberg} games, also known as leader-follower games. In such games the defender acts first, for example by placing their defence system, and the attacker follows with his action, possibly after obtaining information about the placed defence system. In recent years Stackelberg games have seen use in, among others, LAX airport,~\cite{jain2012overview} and the US Coast Guard~\cite{An11}. These games aim to compute randomized policies for the defender to protect target locations from an attacker. Extensions of this model~\cite{Basilico12} have been proposed to generate infinite-horizon patrolling strategies either for mobile resources alone or in concert with static alarm triggers~\cite{basilico2016security,Munoz13}. However, these models cannot be used to reason about the uncertain set of possible  locations of dynamic threats.

Our objective in this work  is not to just compute a patrolling strategy for the mobile sensors, but also to quantify the sensor network's knowledge of the possible locations of active threats and use this information to synthesize strategies for the mobile sensors that provide knowledge guarantees on the threat location over an infinite-time horizon.

As a motivating case study we consider the use of autonomous drones working in cooperation with static sensors in wildlife conservation. UAVs are increasingly being adopted for monitoring of illegal hunting and poaching~\cite{schiffman2014drones}, though they are mostly remotely controlled~\cite{mulero2014remotely}. In Kenya, for example, remotely controlled drones were deployed in 2014 in an attempt to reduce poaching  by providing constant surveillance~\cite{Kenya}, allowing authorities to arrest rhino poachers when they are sensed by the drones. Autonomous UAVs  have not been used in this setting yet, and proposed plans involve drones following pre-programmed paths~\cite{Koh12}. In this paper, we propose a method for automatically constructing  \emph{autonomous reactive surveillance} strategies for multiple mobile sensors (like UAVs) working in concert with static sensors in the field. 

We study the problem of synthesizing strategies for enforcing \emph{temporal surveillance objectives}, such as the requirement to never let the sensor network's uncertainty about the target's location exceed a given threshold, or recapturing the target every time it escapes. To this end, we consider surveillance objectives specified in linear temporal logic (LTL), equipped with basic surveillance predicates. Our computational model is that of a two-player game played on a finite graph, whose nodes represent the joint possible locations of all the mobile sensors and the target, and whose edges model the possible (deterministic) moves between locations. The mobile sensors play the game with partial information, as they can only observe the target when  it is in the area of sight of one of the sensors. The target, on the other hand, always has full information about the locations of all sensors in the network. In that way, we consider a model with one-sided partial information, making the computed strategy for the agent robust against a potentially more powerful adversary.

We formulate surveillance strategy synthesis as the problem of computing a joint winning strategy for the multiple mobile sensors in a partial-information game with a surveillance objective. Partial-information games with LTL objectives have been well studied~\cite{DoyenR11,Chatterjee2013} and it is well known that the synthesis problem is EXPTIME-hard~\cite{Reif84,BerwangerD08}. In a companion publication at CDC 2018 we describe a framework for formalizing \emph{single-agent} surveillance synthesis as a two-player game with partial information, and propose an abstraction-based method for solving such games. The interested reader is referred to the extended version~\cite{arxiv} for details about the abstraction-based synthesis method. The price of resorting to abstraction is the potential overapproximation of the set of possible target locations (that is, loss of precision in the sensors' knowledge) which may make satisfying the surveillance requirements more difficult. There is thus a trade-off between the strictness of the surveillance requirements, i.e, how closely a target needs to be tracked, and the size of the abstract game necessary for a surveillance strategy to exist. 

Sensor networks consisting of a large number of dynamic sensors, as well as static sensors, can achieve better coverage, and thus, in general, can make do with much coarser abstractions to satisfy a given surveillance objective. However, even when using abstraction, the size of the game is exponential in the number of sensors. To address the blow-up of the state space incurred by a large number of sensors, we propose a \emph{decentralized} synthesis method that aims to compute a surveillance strategy for each mobile sensor separately. 

{\bf Contribution:} Our contribution is as follows.
We decompose the original surveillance game into a set of subgames, one for each sensor. Accordingly, the global surveillance objective is  broken up into a local objective for each subgame. Our reduction guarantees that if the local strategy in each subgame satisfies the local surveillance objective, then the composition of the strategies fulfills the global surveillance objective. This allows us to solve each subgame under its local surveillance objective independently, using off-the-shelf reactive synthesis tools. 

 There has been work in decentralized synthesis for GR(1) specifications, however, the synthesis process often involves a centralized computation as in \cite{Kloetzer06} or synchronization \cite{Salar17,Kloetzer11}. Our approach, on the other hand is fully  decentralized and the sensors require no coordination as simply satisfying their local properties guarantees the global objective.


\section{MOTIVATING CASE STUDY}\label{sec:casestudy}
We first describe the multi-agent surveillance synthesis problem informally, in the context of a motivating case study. 

We consider wildlife conservation in Africa, in particular, at the Selous Game Reserve (SGR) located in Tanzania, where the African Black Rhinoceros population is under serious threat due to poaching. We are motivated by a recommended anti-poaching initiative in the SGR by the World Heritage Centre,  to study the use of a sensor network for tracking the position of a potential poacher  with user-specified precision. Since the SGR is a very large area, the network consists of both mobile and static sensors. We apply the  distributed synthesis method proposed in this paper to synthesize surveillance strategies for the mobile sensors that satisfy the desired tracking requirement.

\begin{figure}
\centering
\subfloat[SGR interior landscape. \cite{UN13} \label{fig:SGR-map}]{
\includegraphics[scale=0.15]{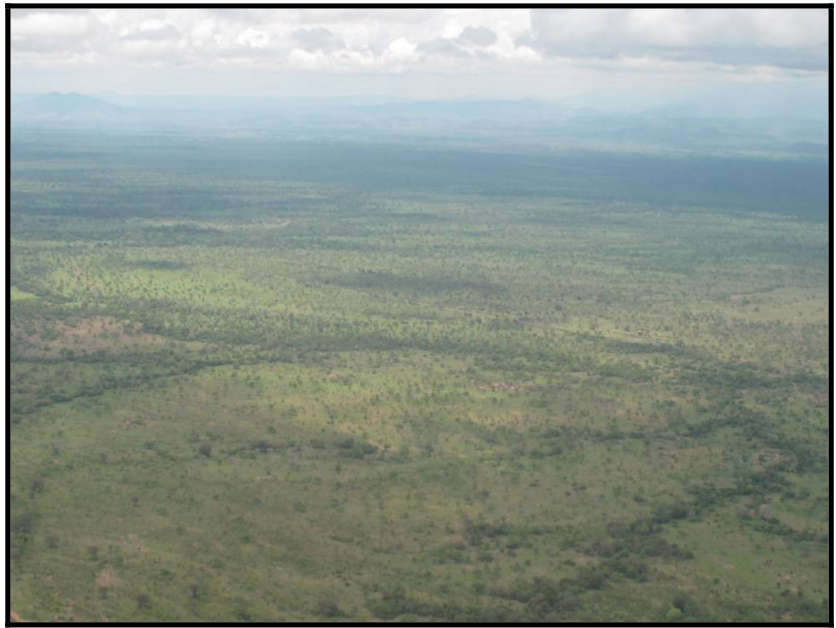}
}\hspace{1cm}
\subfloat[Grid representation of the landscape in \ref{fig:SGR-map}. \label{fig:SGR-grid}]{
\includegraphics[scale=0.15]{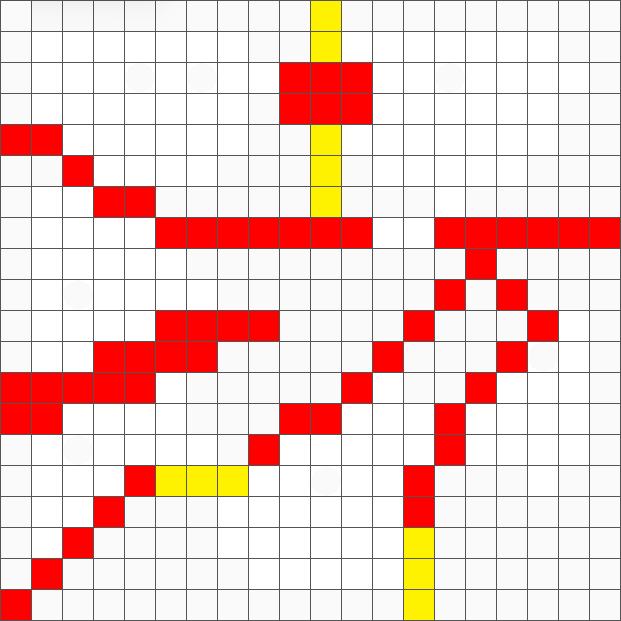}
}

\caption{The landscape in \ref{fig:SGR-map} is coarsely represented as the gridworld in \ref{fig:SGR-grid}. The red regions represent impassable terrain. The yellow areas are the ones covered  by static sensors.}\vspace{-0.5cm}
\label{fig:casestudy}
\end{figure}

Figure \ref{fig:casestudy} shows a section of the SGR that we represent as a gridworld which will form the state space of the game. Each static sensor monitors a given area of the grid (shown in yellow) and detects any presence of the target (i.e., threat) in these states, but cannot determine the target's exact location. The  requirement is to ensure that over and over again, the set of potential locations of the target is reduced to $5$ cells. In other words, every time the target escapes from the vision of all  sensors, the network guarantees that eventually the uncertainty about its position will be reduced to $5$ grid cells.


\section{GAMES WITH SURVEILLANCE OBJECTIVES}
We begin by providing a formal model for describing multi-agent surveillance strategy synthesis problems, in the form of a two-player game between the mobile sensors in a network and a target, in which the sensors have partial information about the target's location. 

\subsection{Multi-Agent Surveillance Game Structures}\label{sec:surveillance-games}
We define a \emph{multi-agent surveillance game structure} to be a tuple $G  = (\states,s^\init,\trans,\vis_1,\dots,\vis_n)$ where:
\begin{itemize}
\item $\states = L_s \times L_t$ is the set of states, where $L_{s} = L_1 \times L_2 \times\dots \times L_n$ is the set of joint locations of the $n$ mobile sensors, $L_i$ is the set of possible locations of sensor $i$,  and $L_t$ is the set of possible locations of the target;
\item $s^\init = (l^{\init}_1,\ldots,l^{\init}_n,l_t^\init)$ is the initial state;
\item $\trans \subseteq \states \times \states$ is the transition relation describing the possible joint moves of the sensors and the target; and
\item  $\vis_1,\dots,\vis_n$ are the \textit{visibility functions} for the $n$ sensors, where $\vis_i: L_{i} \times L_t \to \bools$ maps a state $(l_{i},l_t)$ to $\true$ iff \emph{ position $l_t$ is in the area of sight of $l_i$}.
\end{itemize}

Additionally, we define the \emph{joint visibility function} $\Vis : \states \to \bools$ that maps a state $(l,l_t)$ to $\true$ iff the set $\mathcal{I} = \{i \mid \vis_i(l_i,l_t) = \true\}$ is non-empty. Informally, $\Vis(l_s,l_t)$ is $\true$ if the target is in view of \emph{at least one} of the sensors.

\begin{figure}
\subfloat[Surveillance arena \label{fig:simple-transitions}]{
\begin{tikzpicture}[scale=0.8]
\draw[step=0.5cm,color=gray] (-1.5,-1.5) grid (1,1);
\filldraw[fill=blue,draw=black] (+0.75,+0.75) circle (0.2cm);
\filldraw[fill=red,draw=black] (0,0) rectangle (-0.5,-0.5);
\filldraw[fill=red,draw=black] (-0.5,0) rectangle (-1,-0.5);
\filldraw[fill=red,draw=black] (0,0) rectangle (0.5,-0.5);
\filldraw[fill=gray!80!white,draw=black] (-0.5,0.5) rectangle (-1,1);
\filldraw[fill=gray!80!white,draw=black] (-0.5,0) rectangle (-1,0.5);
\filldraw[fill=gray!80!white,draw=black] (-1,0.5) rectangle (-1.5,1);
\filldraw[fill=gray!80!white,draw=black] (-1,0) rectangle (-1.5,0.5);
\filldraw[fill=gray!80!white,draw=black] (0.5,-0.5) rectangle (1,-1);
\filldraw[fill=gray!80!white,draw=black] (0,-0.5) rectangle (0.5,-1);
\filldraw[fill=gray!80!white,draw=black] (0,-1) rectangle (0.5,-1.5);
\filldraw[fill=gray!80!white,draw=black] (0.5,-1) rectangle (1,-1.5);

\filldraw[fill=blue!40!white,draw=black] (+0.75,+0.75) circle (0.2cm);
\filldraw[fill=orange!40!white,draw=black] (0.25,-0.75) circle (0.2cm);
\filldraw[fill=green!40!white,draw=black]  (-1.27,-1.25) circle (0.2cm);
\node at (-1.25,+0.75) {\tiny{0}};
\node at (-0.80,+0.75) {\tiny{1}};
\node at (-0.30,+0.75) {\tiny{2}};
\node at (0.20,+0.75) {\tiny{3}};
\node at (0.73,+0.75) {\tiny{4}};
\node at (-1.33,+0.25) {\tiny{5}};
\node at (-0.85,+0.25) {\tiny{6}};
\node at (-0.35,+0.25) {\tiny{7}};
\node at (0.25,+0.25) {\tiny{8}};
\node at (0.75,+0.25) {\tiny{9}};
\node at (-1.28,-0.27) {\tiny{10}};
\node at (-0.78,-0.27) {\tiny{11}};
\node at (-0.28,-0.27) {\tiny{12}};
\node at (0.28,-0.27) {\tiny{13}};
\node at (0.75,-0.25) {\tiny{14}};
\node at (-1.3,-0.75) {\tiny{15}};
\node at (-0.8,-0.75) {\tiny{16}};
\node at (-0.3,-0.75) {\tiny{17}};
\node at (0.25,-0.75) {\tiny{18}};
\node at (0.75,-0.75) {\tiny{19}};
\node at (-1.27,-1.25) {\tiny{20}};
\node at (-0.8,-1.25) {\tiny{21}};
\node at (-0.3,-1.25) {\tiny{22}};
\node at (0.25,-1.25) {\tiny{23}};
\node at (0.75,-1.25) {\tiny{24}};
\end{tikzpicture}
\hspace{.1cm}}
\subfloat[Some possible transitions from the initial state in the belief-set game from Example~\ref{ex:simple-belief-game}. Note that, since the set of static sensors is empty, it is omitted from the states. For the sake of readability, some transitions are excluded.\label{fig:simple-belief-game}]{
\input{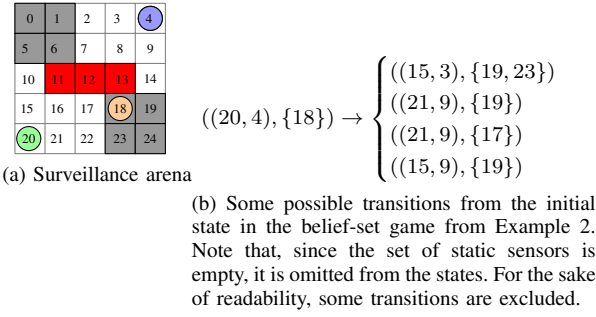}
}
\caption{A simple surveillance game on a grid arena. Obstacles are shown in red. There are two sensors (at locations 20 and 4) coloured in blue and green respectively and the target (at location 18) is orange. The grey states are not visible to either sensor, i.e, $\Vis((20,4,l_t)) = \false$ for all grey $l_t$.}
\label{fig:simple-surveillance-game}
\vspace{-.7cm}
\end{figure}

The transition relation $T$ encodes the one-step move of the target and the $n$ sensors: First, the target makes a move, and then, the sensors move jointly in a synchronized manner. 

We denote with $T{\downarrow}i$ the projection of the transition relation $T$ on the sets of locations of the target and the sensor with index $i$. Formally, we define
$T{\downarrow }i = \{((l_i,l_t),(l_i',l_t')) \in (L_i \times L_t)^2 \mid \exists l_1,l_1',\ldots,l_{i-1},l_{i-1}',l_{i+1},l_{i+1}',\ldots,l_n,l_n' : ((l_1,\ldots,l_n,l_t),(l_1',\ldots,l_n',l_t')) \in T\}.$ 

For a state $(l_s,l_t) \in \states$ we define $\succs_t(l_s,l_t)$ to be the set of possible successor locations of the target:

$\succs_t(l_s,l_{t}) = \{l_{t}' \in L_t \mid \exists l_s'.\ ((l_s,l_t),(l_s',l_t')) \in T\}$.

We extend $\succs_t$ to sets of locations of the target by stipulating that for $L \subseteq L_t$, the set $\post(l_s,L)$ consists of all possible successor locations of the target for states in $\{l_s\} \times L$. Formally, let $\post(l_s, L) = \bigcup_{l_t \in L}\succs_t(l_s,l_t)$.

For a state $(l_s,l_t)$ and a successor location of the target $l_t'$, we denote with $\succs(l_s,l_t,l_t')$ the set of successor locations of the sensors, given that the target moves to $l_t'$: 

$\succs(l_s,l_t,l_t') = \{l_s' \in L_s \mid  ((l_s,l_t),(l_s',l_t')) \in T\}$.

We assume that, for every state $s\in \states$, there exists a state $s' \in \states$ such that $(s,s') \in T$, that is, from every state there is at least one move possible (including self transitions). We also assume, that when the target moves to an invisible location, its position does not influence the possible one-step moves of the sensors. Formally, we require that if $\Vis(l_s,l_t') = \Vis(l_s,{\widehat l_t}')=\false$, then $\succs(l_s,l_t,l_t') = \succs(l_s,{\widehat l_t},\widehat{l_t}')$ for all $l_t,l_t',\widehat l_t,\widehat l_t' \in L_t$. This assumption is natural in the setting where each of the  sensors can move in one step only to locations that are in its sight.

\begin{example}\label{ex:simple-surveillance-game}
Figure~\ref{fig:simple-surveillance-game} shows an example of a multi-agent surveillance game on a grid.  The sets of possible locations $L_i$ and $L_t$ for the each of the sensors and for the target consist of the squares of the  grid. The transition relation $T$ encodes the possible one-step moves of all the sensors and the target on the grid, and incorporates all desired constraints. For example, moving to a location occupied by another sensor or the target, or to an obstacle, is not allowed.
In this example, the function $\vis_i$ encodes straight-line visibility with a range of 2: a location $l_t$ is visible to sensor $i$ from location $l_i$ if there is no obstacle on the straight line between them and the distance between the target and sensor $i$ is not larger than 2. Initially the target is not in the area of sight of the sensors, but the initial position of the target is known. However, once the target moves to one of the locations reachable in one step, in this case, locations $17,19 \text{ and } 23$, this might no longer be the case. More precisely, if the target moves to location $17$, then the green sensor observes its location, but if it moves to one of the other locations, then neither sensor can observe it, and its exact location will not be known. \qed
\end{example}

\subsection{Static Sensors}
We now describe a way to incorporate static sensors in the multi-agent surveillance game framework. Let $G$ be a multi-agent surveillance game structure as defined previously.  

We identify a \emph{static sensor} with a \emph{set of locations $\Lambda \subseteq L_t$ over which it operates}. 
A surveillance game can have multiple static sensors (or none). Let $\mathcal{M} = \{\Lambda_1,\dots,\Lambda_m\}$ be a given set of $m$ static sensors for $G$. For each location $l_t \in L_t$ we define  $J(l_t)$ to be the set of all indices of static sensors such that $l_t$ belongs to the corresponding set of locations, i.e, $J(l_t) = \{j \in \{1,\ldots,m\}\mid l_t\in \Lambda_j\}$. We refer to $J(l_t)$ as the set of \emph{triggered static sensors} at location $l_t$. We also define $J(L) = \bigcup_{l_t \in L} J(l_t)$ for a set of locations $L \subseteq L_t$. 

We assume that sensors do not suffer from  false positives or negatives (studying these is an avenue for future work). 

\subsection{Belief-Set Game Structures}\label{sec:belief-gs}

In surveillance strategy synthesis, we need to state properties of, and reason about, the information which the sensors have, i.e, the \emph{belief} about the location of the target. To this end, we can employ a powerset construction which is commonly used to transform a partial-information game into a perfect-information one, by explicitly tracking the joint knowledge of the sensors as a set of possible locations of the target. In that way we define a two-player game in which one player represents the whole sensor network, and the other player represents the target.

Given a set $A$, we denote with $\mathcal{P}(A) = \{A' \mid A'\subseteq A\}$ the powerset (set of all subsets) of $A$.

Given a  multi-agent surveillance game structure $G  = (\states,s^\init,\trans,\vis_1,\ldots,\vis_n)$ with $m$ static sensors $\mathcal{M} = \{\Lambda_1,\dots,\Lambda_m\}$, we define the corresponding \emph{belief-set game structure} $G_\belief  = (\states_\belief,s^\init_\belief,\trans_\belief)$ where:
\begin{itemize}
\item $\states_\belief = L_s \times \beliefs\times \mathcal{P}(\{1,\ldots,m\})$ is the set of states, where $L_s$ is the set of joint locations of the sensors, and $\beliefs$ the set of \emph{belief sets} describing information about the location of the target, and $\mathcal{P}(\{1,\ldots,m\})$ is the set of possible sets of triggered sensors;
\item $s^\init_\belief = (l^\init_1,\ldots,l^\init_n,\{l_t^\init\},J(l_t^\init))$ is  initial state;
\item $\trans_\belief \subseteq \states_\belief \times \states_\belief$ is the transition relation where $((l_s, B_t,J),(l_s', B_t',J')) \in \trans_\belief$ iff $l_s' \in  \succs(l_s,l_t,l_t')$ for some $l_t \in B_t$ and $l_t' \in B_t'$ , $J' \subseteq J(B_t')$, and one of the following three conditions is satisfied:
\begin{itemize}
\item[(1)] $B_t' = \{l_t'\}$, $l_t' \in \post(l_s,B_t)$, $\Vis(l_s,l_t') = \true$;
\item[(2)] $B_t' = \{l_t' \in \post(l_s,B_t)  \mid  \Vis(l_s,l_t') = \false \} \cap \\
\phantom{B_t' = }\bigcap_{j\in J'}\Lambda_j, \text{ and } J' \neq \emptyset $;
\item[(3)] $B_t' = \{l_t' \in \post(l_s,B_t ) \mid  \Vis(l_s,l_t') = \false\}\setminus\phantom{B_t' = }\bigcup_{j=1}^m \Lambda_j$, and $J' = \emptyset$.
\end{itemize}
\end{itemize}
Condition (1) captures the successor locations of the target that can be observed from one of the mobile sensors' current locations. Condition (2) captures the cases when the target moves to a location that cannot be observed by the mobile sensors, but triggers a non-empty set $J'$ of static sensors. Finally, condition (3) corresponds to the successor locations of the target not visible from the current location of any of the mobile sensors, and not triggering any static sensors.

\begin{example}\label{ex:simple-belief-game}
Consider the surveillance game structure from Example~\ref{ex:simple-surveillance-game}. The initial belief set is $\{18\}$, as the target's initial position is known. Figure~\ref{fig:simple-belief-game} shows some of the successor states of the 
 state $((20,4),\{18\})$  in  $G_\belief$.\qed
\end{example}

Based on  $T_\belief$, we can define the functions $\succs_t : \states_\belief \to \mathcal{P}(\beliefs \times  \mathcal{P}(\{1,\ldots,m\}))$ and  $\succs : \states_\belief \times \beliefs \times  \mathcal{P}(\{1,\ldots,m\}) \to \mathcal{P}(L_s)$ similarly to the corresponding functions defined for $G$. 

A \emph{run} in $G_\belief$ is an infinite sequence $s_0,s_1,\ldots$ of states in $\states_\belief$, where $s_0 = s_\belief^\init$ and $(s_i,s_{i+1}) \in T_\belief$ for all $i$. 

A \emph{strategy for the target in $G_\belief$} is a function $f_t: \states_\belief^+ \to \beliefs \times \mathcal{P}(\{1,\ldots,m\})$ such that $f_t(\pi\cdot s) = (B_t, J)$ implies $(B_t ,J)\in \succs_t(s)$  for every $\pi \in \states_\belief^*$ and $s \in \states_\belief$. That is, a strategy for the target suggests a move resulting in some belief set reachable from a location in the current belief, and a set of triggered sensors.

A \emph{joint strategy for the sensors in $G_\belief$} is a function $f_s : S_\belief^+ \times \beliefs \to S_\belief$ such that, if, $f_s(\pi\cdot s,B_t,J) = (l_s',B_t,J')$ then, $B_t' = B_t$, $J' = J$, and $l_s' \in \succs(s,B_t)$ for every $\pi \in \states_\belief^*$, $s \in \states_\belief$ and $B_t \in \beliefs$. Intuitively, a strategy for the sensors suggests a joint move based on the observed history of the play, the current belief about the target's position, and the set of currently triggered sensors.

The outcome of given strategies $f_s$ and $f_t$ for the sensors and the target in $G_\belief$, denoted $\outcome(G_\belief,f_s,f_t)$, is a run $s_0,s_1,\ldots$ of $G_\belief$ such that for every $i \geq 0$, we have $s_{i+1} = f_s(s_0,\ldots,s_i,B_t^i)$, where $B_t^i = f_t(s_0,\ldots,s_i)$.

\subsection{Temporal Surveillance Objectives}
We consider a set of \emph{surveillance predicates} $\SP = \{p_b \mid b \in \nats_{>0}\}$, where for $b \in \nats_{>0}$ we say that a state $(l_s,B_t)$ in the belief game structure satisfies $p_b$ (denoted $(l_s,B_t) \models p_b$) iff 
$|\{l_t \in B_t  \mid \Vis(l_s,l_t)  = \false \}| \leq b$.
Intuitively, $p_b$ is satisfied by the states in the belief game structure where the size of the belief set does not exceed the threshold $b \in \nats_{>0}$.

We study surveillance objectives expressed in a fragment  of linear temporal logic (LTL) over surveillance predicates.  We consider  \emph{safety surveillance objectives} expressed using the temporal operator $\LTLglobally$ and \emph{liveness surveillance objectives} expressed using the temporal operators $\LTLglobally$ and $\LTLeventually$.

A \emph{safety surveillance objective} $\LTLglobally p_b$ requires that the size of the belief-set never exceeds the given threshold $b$. More formally, an infinite sequence of states $s_0,s_1,\ldots$ in $G_\belief$ satisfies the safety property $\LTLglobally p_b$ if and only if for every $i\geq 0$ it holds that $s_i \models p_b$.  A \emph{liveness surveillance objective} $\LTLglobally\LTLfinally p_b$, on the other hand, requires that 
the size of the belief is smaller or equal to the bound $b$  infinitely often. That is, $s_0,s_1,\ldots$ in $G_\belief$ satisfies $\LTLglobally\LTLfinally p_b$ if for every $i \geq 0$ there exists $j \geq i$ such that $s_j\models p_b$. 

In this paper we consider safety and liveness surveillance objectives, as well as conjunctions of such objectives. We remark the following equivalences of surveillance objectives:
\begin{itemize}
\item $\LTLglobally p_a \wedge \LTLglobally p_b \equiv \LTLglobally p_{\min{(a,b)}}$;
\item $\LTLglobally\LTLfinally p_a \wedge \LTLglobally\LTLfinally p_b \equiv \LTLglobally\LTLfinally p_{\min{(a,b)}}$;
\item if $a \leq b$, then $\LTLglobally p_a \wedge \LTLglobally\LTLfinally p_b \equiv\LTLglobally p_a$. 
\end{itemize}
Using these equivalences, we can restrict our attention to surveillance objectives of one the following  forms: $\LTLglobally p_b$, $\LTLglobally\LTLfinally p_b$ or $\LTLglobally p_a \wedge \LTLglobally\LTLfinally p_b$, where $a > b$.




\subsection{Multi-Agent Surveillance Synthesis Problem}
A \emph{multi-agent surveillance game} is a triple $(G,\mathcal M,\varphi)$, where $G$ is a surveillance game structure, $\mathcal M$ is a set of static sensors,  and $\varphi$ is a surveillance objective. A \emph{winning strategy for the sensors for $(G,\mathcal M,\varphi)$} is a joint strategy $f_s$ for the sensors in the corresponding belief-set game structure $G_\belief$ such that for every strategy $f_t$ for the target in $G_\belief$ it holds that $\outcome(G_\belief,f_s,f_t) \models \varphi$. Analogously, a \emph{winning strategy for the target for $(G,\mathcal M,\varphi)$} is a strategy $f_t$ such that, for every strategy $f_s$ for the mobile sensors in $G_\belief$, it holds that $\outcome(G_\belief,f_s,f_t) \not\models \varphi$.

{\textit{\textbf{Problem statement: }}} Given a multi-agent surveillance game $(G,\mathcal M,\varphi)$,  compute a joint strategy for the mobile sensors that is winning for the game $(G,\mathcal M,\varphi)$.

In the remainder of the paper we show how to solve the multi-agent surveillance synthesis problem in a compositional manner.  The key idea is to decompose the problem into a set of single-sensor surveillance games over smaller sets of locations, and solve each of these games separately.



\section{DISTRIBUTED SURVEILLANCE GAMES}
In the sequel we assume that $L_1 = L_2 = \dots = L_t \triangleq L$  in the surveillance game structure, i.e, all $n$ sensors and the target operate in the same state space. For the remainder of the paper, let $G=(\states,s^\init,\trans,\vis_1,\ldots,\vis_n)$ be a multi-agent surveillance game structure  defined over  $L$, and let $\mathcal M = \{\Lambda_1,\dots,\Lambda_m\}$ be a set of static sensors. We define a  \emph{state-space partition} of size $n$ of the set $L$ of locations in a game structure $G$ to be a tuple $\widetilde L =  (\widetilde L_1,\ldots,\widetilde L_n)$ of subsets of $L_i$ such that  $\bigcup_{i=1}^n \widetilde L_i = L $, and $\widetilde L_i \cap \widetilde L_j  = \emptyset$ for $i \neq j$.

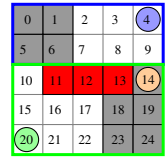
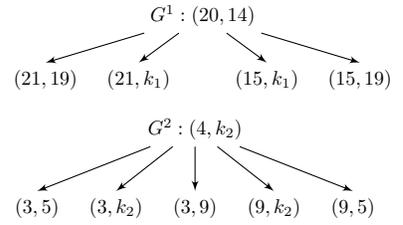
\begin{figure}
\subfloat[Multi-agent surveillance game partitioned into two subgames. \label{part-grid}]{
\begin{tikzpicture}[scale=0.8]
\draw[step=0.5cm,color=gray] (-1.5,-1.5) grid (1,1);
\filldraw[fill=red,draw=black] (0,0) rectangle (-0.5,-0.5);
\filldraw[fill=red,draw=black] (-0.5,0) rectangle (-1,-0.5);
\filldraw[fill=red,draw=black] (0,0) rectangle (0.5,-0.5);
\filldraw[fill=gray!80!white,draw=black] (-0.5,0.5) rectangle (-1,1);
\filldraw[fill=gray!80!white,draw=black] (-0.5,0) rectangle (-1,0.5);
\filldraw[fill=gray!80!white,draw=black] (-1,0.5) rectangle (-1.5,1);
\filldraw[fill=gray!80!white,draw=black] (-1,0) rectangle (-1.5,0.5);
\filldraw[fill=gray!80!white,draw=black] (0.5,-0.5) rectangle (1,-1);
\filldraw[fill=gray!80!white,draw=black] (0,-0.5) rectangle (0.5,-1);
\filldraw[fill=gray!80!white,draw=black] (0,-1) rectangle (0.5,-1.5);
\filldraw[fill=gray!80!white,draw=black] (0.5,-1) rectangle (1,-1.5);
\filldraw[fill=gray!80!white,draw=black] (0.5,-0.5) rectangle (1,0);

\filldraw[fill=blue!40!white,draw=black] (+0.75,+0.75) circle (0.2cm);
\filldraw[fill=orange!40!white,draw=black] (0.75,-0.25) circle (0.2cm);
\filldraw[fill=green!40!white,draw=black]  (-1.27,-1.25) circle (0.2cm);
\filldraw[fill=none,draw=blue,line width=0.4mm] (-1.5,0) rectangle (1,1);
\filldraw[fill=none,draw=green,line width=0.4mm] (-1.5,-1.5) rectangle (1,0);
\node at (-1.25,+0.75) {\tiny{0}};
\node at (-0.80,+0.75) {\tiny{1}};
\node at (-0.30,+0.75) {\tiny{2}};
\node at (0.20,+0.75) {\tiny{3}};
\node at (0.73,+0.75) {\tiny{4}};
\node at (-1.33,+0.25) {\tiny{5}};
\node at (-0.85,+0.25) {\tiny{6}};
\node at (-0.35,+0.25) {\tiny{7}};
\node at (0.25,+0.25) {\tiny{8}};
\node at (0.75,+0.25) {\tiny{9}};
\node at (-1.28,-0.27) {\tiny{10}};
\node at (-0.78,-0.27) {\tiny{11}};
\node at (-0.28,-0.27) {\tiny{12}};
\node at (0.28,-0.27) {\tiny{13}};
\node at (0.75,-0.25) {\tiny{14}};
\node at (-1.3,-0.75) {\tiny{15}};
\node at (-0.8,-0.75) {\tiny{16}};
\node at (-0.3,-0.75) {\tiny{17}};
\node at (0.25,-0.75) {\tiny{18}};
\node at (0.75,-0.75) {\tiny{19}};
\node at (-1.27,-1.25) {\tiny{20}};
\node at (-0.8,-1.25) {\tiny{21}};
\node at (-0.3,-1.25) {\tiny{22}};
\node at (0.25,-1.25) {\tiny{23}};
\node at (0.75,-1.25) {\tiny{24}};

\end{tikzpicture}
\hspace{.3cm}}
\subfloat[Transitions from initial states in subgames. \label{part-grid-trans}]{
\begin{minipage}{5.0cm}
\vspace{-0.8cm}


%

\begin{tikzpicture}[node distance=.9 cm,auto,>=latex',line join=bevel,transform shape,scale=.75]
\node at (0,0) (s0) {$G^1: (20,14)$};
\node  [below left of=s0,yshift=-.5cm,xshift=-0.5cm] (s2) {$(21,k_1)$};
\node  [below right of=s0,yshift=-.5cm,xshift=0.5cm] (s3) {$(15,k_1)$};
\node  [left of=s2,xshift=-0.75cm] (s1) {$(21,19)$};
\node  [right of=s3,xshift=0.75cm] (s4) {$(15,19)$};
\draw [->] (s0) edge (s1.north);
\draw [->] (s0) edge (s2.north);
\draw [->] (s0) edge (s3.north);
\draw [->] (s0) edge (s4.north);
\end{tikzpicture}

\begin{center}
\begin{tikzpicture}[node distance=.9 cm,auto,>=latex',line join=bevel,transform shape,scale=.75]
\node at (0,0) (s0) {$G^2: (4,k_2)$};
\node  [below of=s0,yshift=-.5cm] (s1) {$(3,9)$};
\node  [right of=s1,xshift=0.5cm] (s2) {$(9,k_2)$};
\node  [left of=s1,xshift=-0.5cm] (s3) {$(3,k_2)$};
\node  [left of=s3,xshift=-0.5cm] (s4) {$(3,5)$};
\node  [right of=s2,xshift=0.5cm] (s5) {$(9,5)$};
\draw [->] (s0) edge (s1.north);
\draw [->] (s0) edge (s2.north);
\draw [->] (s0) edge (s3.north);
\draw [->] (s0) edge (s4.north);
\draw [->] (s0) edge (s5.north);
\end{tikzpicture}
\end{center}

\end{minipage}
}
\hspace{0.2cm}
\vspace{-0.3cm}
\caption{Partitioning of the state space of a surveillance game into two subgames with locations $\widetilde{L}_1$ (green) and $\widetilde{L}_2$ (blue). }
\label{fig:simple-dist-game}

\end{figure}

\subsection{Surveillance Subgames}\label{sec:subgames}
We now describe how, given a state-space partition $\widetilde L =  (\widetilde L_1,\ldots,\widetilde L_n)$, to construct a tuple of single-agent surveillance game structures $\widetilde G = (G^1,\ldots,G^n)$ that contains one  \emph{surveillance subgame} $G^i$ for each mobile sensor  $i$. Each subgame, $G^i$ is defined over the subset of locations $\widetilde{L}_i$. Since the target and sensors operate on the same state space we will have $\widetilde{L}^i_s = \widetilde{L}^i_t = \widetilde{L}_i$. Additionally, to each $\widetilde{L}^i_t$ we add an \emph{auxiliary location} $k_i$ that encapsulates all possible locations of the target that are outside of this subgame's region, i.e., all locations in $L \setminus \widetilde{L}_i$.  We then model transitions leaving or entering $\widetilde{L}^i_t$ as transitions to or from location $k_i$ respectively.
We require that the initial location $l_i^\init$ of sensor $i$ is in $\widetilde{L}_i$.

Formally, given a subset  $\widetilde{L}_i \subseteq L$ we define the \emph{subgame} of $G$ corresponding to sensor $i$ as the tuple $G^i = (\widetilde{S}_i,\widetilde{s}_i^{\init},\widetilde{T}_i,\widetilde{\vis}_i)$  where:
\begin{itemize}
\item $\widetilde{S}_i= \widetilde{L}_i \times (\widetilde{L}^i_t \cup k_i)$ is the set of states.
\item $\widetilde s_i^\init = (l_i^\init,\widetilde l_t)$ is the initial state, where $\widetilde l_t =  l_t^\init$, if
$ l_t^\init \in \widetilde L_i$, and $\widetilde l_t = k_i$ otherwise.
\item The set $\widetilde{T}_i$ consists of two types of transitions: the transitions in $T{\downarrow } i$ that originate and end in the subgame's region are preserved as they are. Transitions of the target exiting or entering $\widetilde{L}^i_t$ are replaced by transitions to and from location $k_i$ respectively, since $k_i$ represents all target locations outside of  $\widetilde{L}^i_t$. 
Formally, for every pair of states $(\widetilde{l}_i,\widetilde{l}_t) \in \widetilde{S}_i$ and $(\widetilde{l}_i',\widetilde{l}_t') \in \widetilde{S}_i$ we have that $((\widetilde{l}_i,\widetilde l_t),(\widetilde{l}_i',\widetilde l_t')) \in \widetilde T_i$ if and only if there exists a transition
 $((\widetilde{l}_i,l_t),(\widetilde{l}_i',l_t')) \in T{\downarrow}i$ for which the following conditions are satisfied:
 \begin{itemize}
 \item if $\widetilde l_t \in \widetilde L_t ^i$ and $\widetilde l_t' \in \widetilde L_t ^i$, then 
 $\widetilde l_t = l_t$ and $\widetilde l_t'= l_t'$, that is, we have a \emph{transition internal for the region $\widetilde L_t^i$};
 \item if $\widetilde l_t \in \widetilde L_t ^i$ and $\widetilde l_t' =  k_i$, then 
 $l_t \in \widetilde L_t^i$ and $l_t' \not\in \widetilde L_t^i$, that is, we have a \emph{transition exiting the region $\widetilde L_t^i$}; 
 \item if $\widetilde l_t= k_i$ and $\widetilde l_t' \in  \widetilde L_t ^i$, then 
 $l_t \not \in \widetilde L_t^i$ and $l_t' \in \widetilde L_t^i$, that is, we have a \emph{transition entering the region $\widetilde L_t^i$}; 
 \item if $\widetilde l_t= k_i$ and $\widetilde l_t' =  k_i$, then 
 $l_t \not \in \widetilde L_t^i$ and $l_t' \not\in \widetilde L_t^i$, that is, we have a \emph{transition completely outside $\widetilde L_t^i$}.
\end{itemize}  

  \item The visibility function $\widetilde{\vis}_i$ in the subgame $G^i$ agrees with the visibility function $\vis_i$ of sensor $i$ in the original game when the target's location is in the subgame's region. Target locations outside of the region $\widetilde L_t^i$  (summarized by location $k_i$) are invisible to the sensor in the subgame. Formally, $\widetilde{\vis}_i(\widetilde{l}_i,l_t) = \vis_i(\widetilde{l}_i,l_t)$  when $l_t \in \widetilde{L}_t^i$, and $\widetilde{\vis}_i(\widetilde{l}_i,l_t) =\false$ if $l_t  = k_i$.
\end{itemize}
\begin{example}
In Figure \ref{part-grid}, we have two subgames: $G^1$ for the green mobile sensor and $G^2$ for the blue one. The initial states in the subgames are $s_1 = (20,14)$, and $s_2 = (4,k_2)$. Recall that $k_i$ is an indicator state to represent that the target is not subgame $i$. The transitions shown in Figure \ref{part-grid-trans} show that the target has the ability to leave $G^1$ and enter $G^2$. 
\qed
\end{example}

Note that in this construction, sensor $i$ is not able to leave the region of locations $\widetilde{L}_i$. Furthermore, all the information about the target's behaviour outside of  the subgame's region is completely hidden from the mobile sensor controller, since all locations outside of  $\widetilde{L}_t^i$ are represented by the single location $k_i$.
In section \ref{sec:local-games}, we discuss the local knowledge (belief)  of sensor $i$ in the game structure $G^i$.
 
\subsection{Static Sensors in Subgames}
We assumed that all information about the target's behaviour outside of subgame $G^i$ is completely  hidden from sensor $i$. Hence, sensor $i$ is only privy to static sensors that operate in the state space of the subgame $G^i$, i.e, static sensors $\Lambda_m$ where $\Lambda_m \cap \widetilde{L}_i \neq \emptyset$. For simplicity of the presentation we assume that each static static sensor operates in exactly one region $L_i$. Our results can easily be extended to the general case. We define $Q_i = \{i \mid \Lambda_{i} \cap \widetilde{L}_i \neq \emptyset\}$ to be the set of static sensors operating in the subgame  $G^i$. 

\subsection{Local Beliefs in Surveillance Subgames}\label{sec:local-games}
A surveillance subgame is a game structure with a single mobile sensor and some number of static sensors, and thus, is  a special case of multi-agent surveillance game structure. With this, the definition of  belief-set game structures from Section~\ref{sec:belief-gs} directly applies to surveillance subgames.

In the belief-set game structure for a surveillance subgame $G^{i} = (\widetilde{S}_i,\widetilde{s}_i^{init},\widetilde{T}_i,\widetilde{\vis}_i)$ with static sensors $Q_i$, the belief sets represent the local belief of sensor $i$. More specifically, a belief set in $G^{i}_\belief$ is an element of  $\mathcal{P}(\widetilde{L}^i_t \cup \{k_i\})$, and can thus contain the auxiliary location. Intuitively, if $ k_i$ is present in the sensor's current belief, then the target could possibly be outside of the local set of locations $\widetilde L_i$, or if the belief is the singleton $\{k_i\}$, then sensor $i$ knows for sure that the target is outside of its region. Additionally, if there is a triggered static sensor in the region, the sensor will know that the target must be in the state space of the static sensor and $k_i$ cannot be in the belief. Due to the definition of surveillance subgames in Section~\ref{sec:subgames}, the location $k_i$ \emph{must} be in the belief of sensor $i$ \emph{whenever it is possible that the target is outside of its region}. If $n\geq 2$, then at every given time $k_i$ must be in the belief set of at least one sensor (possibly several).  
We define the \emph{global interpretation} $\llbracket B_t\rrbracket$ of a belief set $B_t$ in $G^{i}_\belief$, which is a set of locations in $G$, as
 \[\llbracket B_t\rrbracket = \begin{cases}
B_t & \text{if } k_i \not\in B_t\\
B_t \cup (L \setminus \widetilde L_i) & \text{if } k_i  \in B_t.
\end{cases}
\]


Strategies of sensor $i$ in the belief-set game $G^{i}_\belief$ depend only on the sequence of states in this game, and thus, only on local information. Following the definitions in Section~\ref{sec:belief-gs}, the outcome of a pair of given strategies $f_{i}$ and $f_{t_i}$ for the sensor and the target in $G^{i}_{\belief}$  is a sequence of  states in  $G^{i}_{\belief}$, each of which is a pair consisting of a location of sensor $i$ and a belief-set for sensor $i$ in $G^{i}_{\belief}$.


\subsection{Distributed Surveillance Synthesis Problem}\label{sec:distributed-problem}
Given a state-space partitioning $\widetilde L$ and the corresponding tuple of subgames $\widetilde G = (G^1,\ldots,G^n)$
we will define a distributed surveillance strategy synthesis problem, which, intuitively, asks to synthesize strategies for the sensors in the individual belief subgames, such that together they guarantee the global surveillance objective.  In this section we formalize this intuitive problem description. We first need to define what it means for the individual sensor strategies to jointly satisfy together a global requirement.

The surveillance requirements are defined in terms of the belief-states in $G_\belief$, but strategies in the belief subgames are defined in terms of sequences of local belief states. Hence, we need to define a mapping of states of the form $((l_1,\ldots,l_n),B_t,J)$ to elements of $\mathcal{P}(\widetilde{L}^i_t \cup \{k_i\})$ for each $i$. Since, by definition, a strategy for sensor $i$ in the corresponding belief subgame guarantees that it remains in $\widetilde L_i$, we only need to define the mapping for states  $l_i \in\widetilde{L}_i$.

Formally, for a state $((l_1,\ldots,l_n),B_t,J)$ in $G_\belief$  we define its projection on belief subgame $i$ as $((l_1,\ldots,l_n),B_t,J){\downarrow}i = (l_i,B_t{\downarrow}i,J \cap Q_i)$, where
\[B_t{\downarrow}i = \begin{cases}
B_t& \text{if }  B_t \subseteq \widetilde L_i, \\
(B_t \cap \widetilde L_i) \cup \{k_i\} & \text{otherwise}.
\end{cases}\]
The mapping extends to sequences of states in the usual way. 

Intuitively, this mapping projects the joint knowledge of the sensors in $G_\belief$ onto the local belief of each sensor, where the sensors do not share their local beliefs with each other, that is, the sensors have no information about the target's position outside of their own region. The global, shared belief of the sensors is formed by the combination of their local beliefs. More precisely, this is the intersection of the  global interpretation of the local beliefs. Indeed, it is easy to see that the property $B_t = \bigcap_{i=1}^n \llbracket B_t{\downarrow}i\rrbracket$ holds.

Now we are ready to define the joint strategy of the sensors in $G_\belief$ obtained by executing together a given set of sensor strategies in the individual subgames. 
Let $f_{s_1},\ldots,f_{s_n}$ be strategies for the sensors in the belief subgames $(G^{1}_{\belief},\ldots,G^{n}_{\belief})$. We define the \emph{composition} $f_{s_1} \otimes\ldots\otimes f_{s_n}$ of $f_{s_1},\ldots,f_{s_n}$, which is a joint strategy $f_s$ for the sensors in $G_\belief$, as follows:
for every sequence $s_0,\ldots,s_k$ of states in $G_\belief$, global belief $B_t \in \beliefs$ and set of triggered sensors $J \subseteq \{1,\ldots,m\}$, we let
\[f_s(s_0,\ldots,s_k,B_t,J) = (l_1,\ldots,l_n),\]
where $l_i = f_{s_i}((s_0,\ldots,s_k){\downarrow}i,B_t{\downarrow}i, J \cap Q_i)$ for each $i$. 

\emph{Remark}. If, for some $i$, the projection $(s_0,\ldots,s_k){\downarrow}i$ is undefined, then $f_s(s_0,\ldots,s_k,B_t)$ is undefined. However, by the definition of each $f_{s_i}$ we are guaranteed that the projection is defined for every prefix consistent with $f_{s_i}$.

Intuitively, the joint strategy $f_{s_1} \otimes\ldots\otimes f_{s_n}$ makes decisions consistent with the choices of the individual strategies $f_{s_1}, \ldots, f_{s_n}$ in the respective belief subgames.


Our goal is to synthesize a joint strategy $f_s$ that enforces a given surveillance property in the belief-set game $G_\belief$ by synthesizing individual strategies for all the sensors in the corresponding belief subgames. That is, we want to solve the following \emph{distributed surveillance synthesis problem}. 
 
{\textit{\textbf{Problem statement: }}}Given a multi-agent surveillance game $(G,\mathcal M,\varphi)$ with $n$ sensors, and a state-space partition $\widetilde{L}$, compute strategies $f_{s_1},\ldots,f_{s_n}$ for the sensors in the belief subgames $G^1_\belief,\ldots,G^n_\belief$ respectively, such that the composed strategy $f_{s_1}\otimes\ldots\otimes f_{s_n}$ is a joint winning strategy for the sensors in the surveillance game $(G,\mathcal M,\varphi)$.

\smallskip

Thus, in the distributed surveillance synthesis problem we have to compute strategies $f_{s_1},\ldots,f_{s_n}$  such that for every strategy $f_t$ for the target in $G_{\belief}$ it holds that $\outcome(G_{\belief},f_{s_1}\otimes\ldots\otimes f_{s_n},f_{t}) \models \varphi$. To this end, we have to provide \emph{local surveillance objectives} for all the sensors, such that if all strategies are winning with respect to their local objectives, then their composition is winning with respect to the original surveillance objective. In this way we will reduce the multi-agent surveillance synthesis problem to $n$ single-agent surveillance problems over smaller sets of locations. This reduction is the subject of the next section.

\section{FROM GLOBAL TO LOCAL SPECIFICATIONS}
\everypar{\looseness=-1}
In order to reduce the multi-agent surveillance synthesis problem for a given surveillance specification $\varphi$ to solving a number of single-sensor surveillance subgames, we need to provide local surveillance objectives for the individual subgames. The local objectives should be such that by composing the strategies that are winning with respect to the local objectives we should obtain a strategy that is winning for the global surveillance objective. More precisely, we have to provide local surveillance specifications $\varphi_1,\ldots,\varphi_n$ such that if for each $i$ it holds that $f_{s_i}$ is  a winning strategy for the sensor in $(G^i,Q_i,\varphi_i)$, then the strategy $f_{s_1}\otimes \ldots \otimes f_{s_n}$ is  a joint winning strategy for the sensors in $(G,\mathcal M,\varphi)$.

Recall that the surveillance objective $\varphi$ is of the form $\LTLglobally p_b$, or $\LTLglobally\LTLfinally p_b$, or $\LTLglobally p_a \wedge \LTLglobally\LTLfinally p_b$, where $a > b$. We will provide translations for each of these types of  specifications.

First, note that the belief sets in a belief subgame $G^i_\belief$ can contain the auxiliary location $k_i$, which represents all locations in $L \setminus \widetilde L_i$. Thus, when the local belief set contains $k_i$, the size of the global belief set depends on the local beliefs of the other agents as well. We have to account for this in the translation from global into local surveillance objectives.

\begin{example}\label{ex:global-local-safety}
Consider the global safety surveillance specification $\LTLglobally p_5$ in a network with two mobile sensors. In this case we can reduce the multi-agent surveillance problem to two single-agent surveillance games, each of which has $\LTLglobally p_3$ as the local specification. 
\everypar{\looseness=-1}
To see why, consider the two possible cases of local belief set of sensor $1$ whose size is less than or equal to $3$. If $k_1$ is not part of the belief set of sensor $1$, then the target is definitely in the region of sensor $1$, meaning that the global belief is of size less than or equal to $3$, and hence smaller that $5$. If, on the other hand, $k_1$ is part of the local belief of sensor $1$, then the target can be in at most $2$ locations in $\widetilde L_1$. If at the same time we have that the local belief of sensor $2$ is of size at most $3$, this would guarantee that the size of the global belief does not exceed $5$. 
\everypar{\looseness=-1}
Local specifications $\LTLglobally p_4$, on the other hand do not imply the global specification. Indeed, if at a given point in time both sensors have local beliefs of size $4$, each of which contains the corresponding location $k_i$, the resulting global belief will be of size $6$ and thus violate the global specification.\qed
\end{example}

Generalizing the observations made in this example, for any number of sensors $n \geq 2$ and global safety surveillance objective $\LTLglobally p_b$, we define the local safety surveillance objective for each of the sensors, denoted $\locspec(\LTLglobally p_b,n)$, as $\locspec(\LTLglobally p_b,n) \triangleq \LTLglobally p_c,$ where $c = \lfloor{\frac{b}{n}}\rfloor+1$. Since $n \geq 2$ and $b >0$, we have $c \leq b$.

Note that this translation is conservative, since if according to the belief of sensor $i$ the target could be outside its region, it should guarantee that the number of locations in its own region the target could be in is at most  $\lfloor{\frac{b}{n}}\rfloor$, even if the target can possibly be in only one of the other regions. This conservativeness is necessary to guarantee soundness in the absence of coordination between the sensors.

We now turn to liveness surveillance objectives. It is easy to see that each sensor guaranteeing a small enough local belief infinitely often is not enough to satisfy the global surveillance objective, since the local guarantees can happen in time-steps different for the different sensors.

\begin{example}\label{ex:global-local-liveness}
Consider the global surveillance specification $\LTLglobally\LTLfinally p_5$ for a network with two sensors. Suppose $f_1$ is a strategy for the sensor in $G^1_\belief$,  which ensures that every even step the size of the local belief is $10$, and every odd step the local belief contains $k_1$ and its size is $3$. Strategy  $f_2$ in $G^2_\belief$,  is similar, but even and odd steps are interchanged: every even step the local belief contains $k_2$ and its size is $3$, and every odd step the size of the local belief is $10$.  Thus, while $f_1$ and $f_2$ guarantee that their local belief is "small enough" infinitely often, they do this at different steps.
\end{example}

We circumvent the problem illustrated in this example by requiring that each sensor satisfies the liveness guarantee on its own. For this, we have to consider two cases. First, if from some point on sensor $i$ always knows that the target is outside of its region, it has no obligation to satisfy the liveness surveillance guarantee. If, on the other hand, according to sensor $i$'s belief the target could be in $\widetilde L_i$ infinitely often (note that this is true for at least one sensor), then it has to satisfy the corresponding liveness guarantee.

In order to capture this intuition, we need two additional types of surveillance predicates. First, we need to be able to express the negation of the property that the local belief of sensor $i$ is the singleton $\{k_i\}$ (which means that sensor $i$ knows that the target is outside $\widetilde L_i$). For this, we introduce the predicate $\mathit{belief} \neq \{k_i\}$. Second, in order to express the local liveness guarantee, we need to be able to state that $k_i$ is not in $\widetilde L_i$ (which means that sensor $i$ knows that the target is in its region). The predicate we introduce for this property is $k_i \not\in\mathit{belief}$. Both predicates can be interpreted over belief sets similarly to $p_b$ and incorporated in LTL.
 
Formally, we define the local liveness specification for sensor $i$ denoted $\locspec_i(\LTLglobally\LTLfinally p_b)$ as
\[
\begin{array}{lll}
\locspec_i(\LTLglobally\LTLfinally p_b) & \triangleq &
\big(\LTLglobally\LTLfinally (\mathit{belief} \neq \{k_i\} )\big)\rightarrow\\&& \big(\LTLglobally\LTLfinally (p_b \wedge (k_i \not\in\mathit{belief}))\big).
\end{array}
\]
Note that the agent cannot trivially satisfy $\locspec_i(\LTLglobally\LTLfinally p_b)$, since the belief set is defined precisely by it's sequence of observations and is not under the agent's direct control.

This translation is again conservative, since it would suffice that the liveness guarantee is satisfied by a single sensor. However, these can be different sensors for different behaviours of the target. Thus, we require that \emph{every} sensor $i$ satisfies $\locspec_i(\LTLglobally\LTLfinally p_b)$. This requires that if the target crosses from one region to another infinitely often, then both sensors have to satisfy the liveness surveillance objective.

Finally, for a global surveillance specification $\LTLglobally p_a \wedge \LTLglobally\LTLfinally p_b$, the local surveillance specification for sensor $i$ is
\[\locspec_i(\LTLglobally p_a\wedge  \LTLglobally\LTLfinally p_b,n)\triangleq
\locspec(\LTLglobally p_a,n) \wedge\locspec_i(\LTLglobally\LTLfinally p_b) .
\]

Slightly abusing the notation, we denote with $\locspec_i(\varphi,n)$ the local surveillance specification for sensor $i$ for any of the three types of global surveillance specifications.

The next theorem, which follows  from the definition of the local specifications, states the soundness of the reduction.

\begin{theorem}\label{thm:soundnes}
Let  $(G,\mathcal M,\varphi)$ be a multi-agent surveillance game with $n$ sensors, where $\varphi$ is of the form $\LTLglobally p_b$, or $\LTLglobally\LTLfinally p_b$, or $\LTLglobally p_a \wedge \LTLglobally\LTLfinally p_b$, where $a > b$. Let $\widetilde{L}$ be a state-space partition. Suppose that $f_1,\ldots,f_n$ are strategies for the sensors in the subgames $G^1_\belief,\ldots,G^n_\belief$ respectively, such that for each sensor $i$ the strategy $f_i$ is winning in the surveillance game $(G^i,Q_i,\locspec_i(\varphi,n))$. Then, it holds that the composed strategy $f_{s_1}\otimes\ldots\otimes f_{s_n}$ is a joint winning strategy for the sensors in the surveillance game $(G,\mathcal M,\varphi)$.
\end{theorem}
%

\section{EXPERIMENTAL EVALUATION}\label{sec:experiments}
We now return to the case study outlined in Section~\ref{sec:casestudy}. We have implemented the proposed method in \texttt{Python}, using the \texttt{slugs} reactive synthesis tool~\cite{EhlersR16}, and evaluated it on the multi-agent surveillance game modelling the problem described in Section~\ref{sec:casestudy}. The experiments were performed on an Intel i5-5300U 2.30 GHz CPU with 8 GB of RAM.

\begin{figure}
	\centering
\subfloat[The gridworld in \ref{fig:SGR-grid}\newline partitioned into 6 subgames. \label{fig:experiment}]{
\includegraphics[scale=0.18]{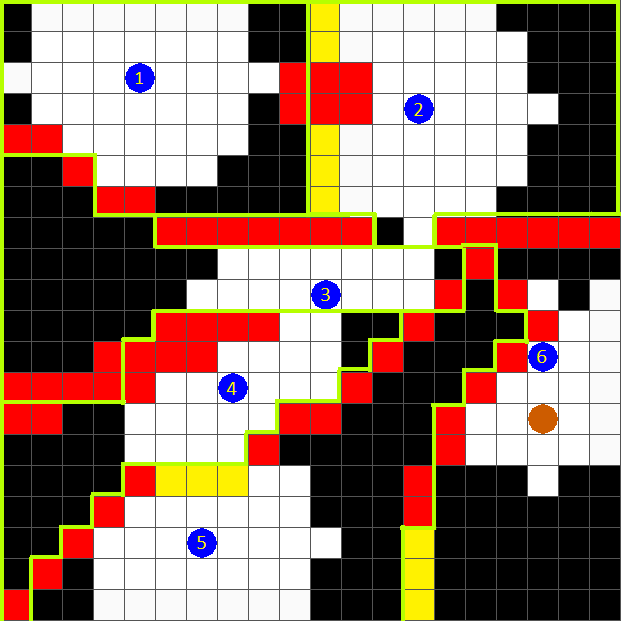}
\hspace{1.5cm}}
\subfloat[The gridworld in \ref{fig:SGR-grid}\newline partitioned into 3 subgames. \label{fig:3experiment}]{
\includegraphics[scale=0.18]{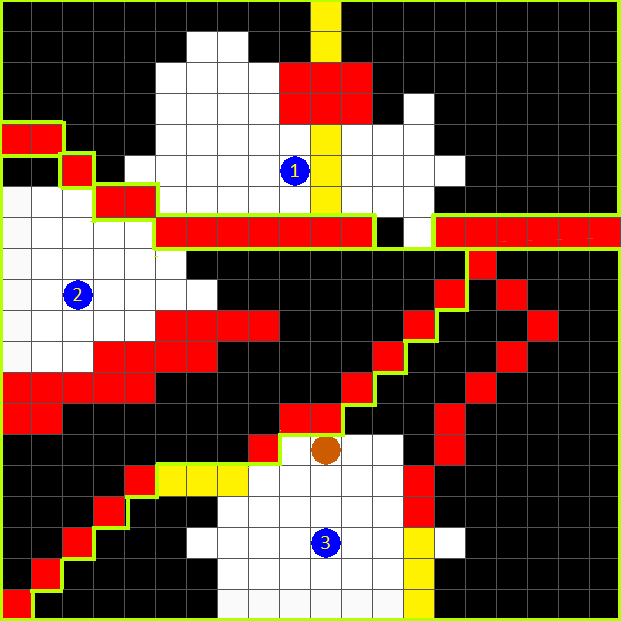}
\hspace{.3cm}}

\caption{Cases with 6 mobile sensors in Fig \ref{fig:experiment} and 3 mobile sensors in Fig \ref{fig:3experiment}. The mobile sensors are blue circles and the target is represented in orange. Yellow regions represent static sensors.The red cells represent impassable terrain (such as dense foliage) that cannot be seen through by the sensors. Black cells are locations not visible to any sensor.}\label{fig:bigexp}\vspace{-0.5cm}
\end{figure}

 We analyzed two scenarios. In Figure \ref{fig:experiment}, we have six \emph{mobile sensors}. We compare the surveillance strategy with the situation in Figure \ref{fig:3experiment} where we have three \emph{mobile sensors}. In both cases there are four \emph{static sensors} depicted in yellow in Figure~\ref{fig:bigexp}. Our global surveillance task is $\LTLsquare \LTLdiamond p_5$, i.e, we need to infinitely often bring the belief of the target location to 5 cells or lower. 

Solving either case centralized is not computationally feasible as the state space grows exponentially with the number of sensors - we will have in the order of $400^6$ and $400^3$ states respectively. Thus, we partition the multi-agent surveillance game into subgames as shown in Figures \ref{fig:experiment} and \ref{fig:3experiment}. We then solve each game individually with local specifications $\locspec_i(\LTLglobally\LTLfinally p_5)$. We solve these \emph{single-agent} surveillance games using an abstraction-based method detailed in a companion publication at CDC 2018, detailed in~\cite{arxiv}. 
We report the synthesis times in Table \ref{tab:synthtime}.

\begin{figure}
	
	\begin{minipage}{5.0cm}
		\centering
		\subfloat[$t_8$ \label{fig:case1t2}]{
			\includegraphics[scale=0.12]{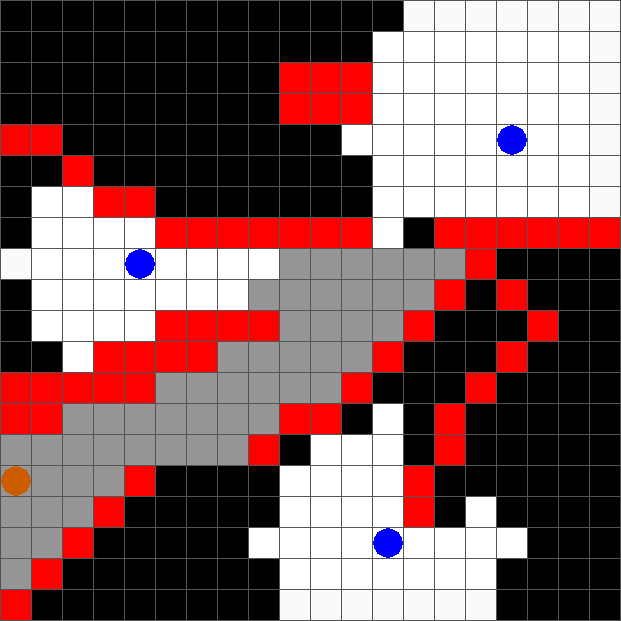}\hspace{.7cm}
		}
		\subfloat[$t_{12}$ \label{fig:case1t3}]{
			\includegraphics[scale=0.12]{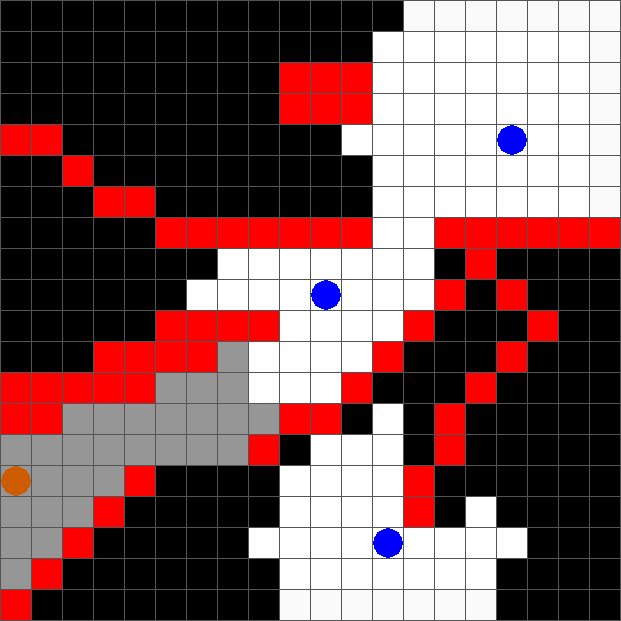}\hspace{.7cm}
		}
		\subfloat[$t_{16}$ \label{fig:case1t4}]{
			\includegraphics[scale=0.12]{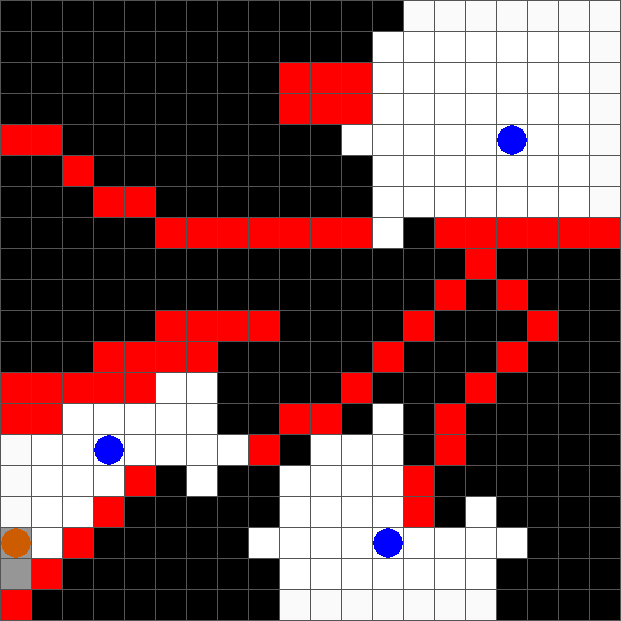}\hspace{.7cm}
		}
	\end{minipage}
	\begin{minipage}{5.0cm}
		\centering
		\subfloat[$t_{18}$  \label{fig:case1t5}]{
			\includegraphics[scale=0.12]{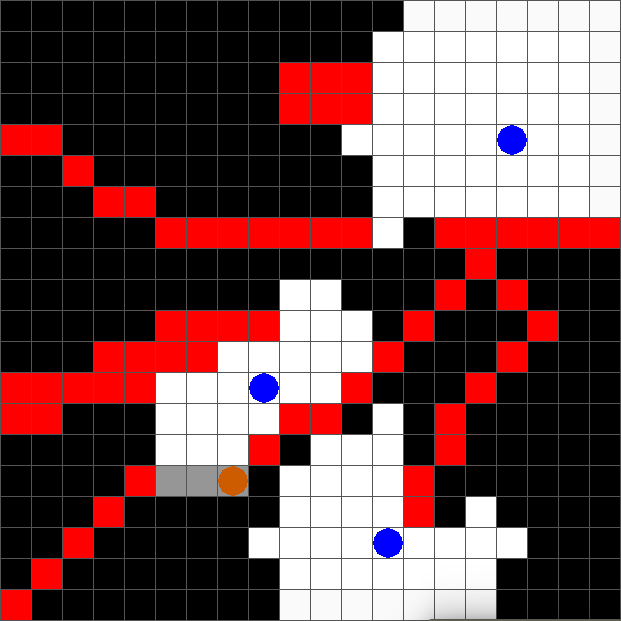}\hspace{.7cm}
		}
		\subfloat[$t_{20}$ \label{fig:case1t6}]{
			\includegraphics[scale=0.12]{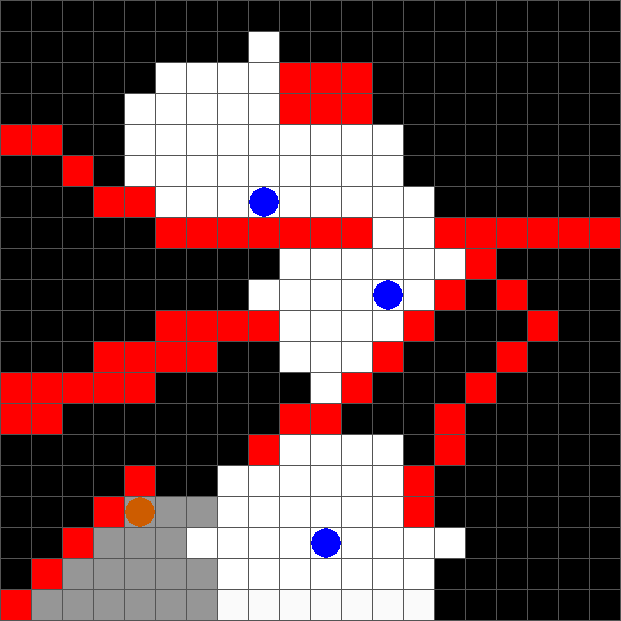}\hspace{.7cm}
		}
		\subfloat[$t_{22}$ \label{fig:case1t7}]{
			\includegraphics[scale=0.12]{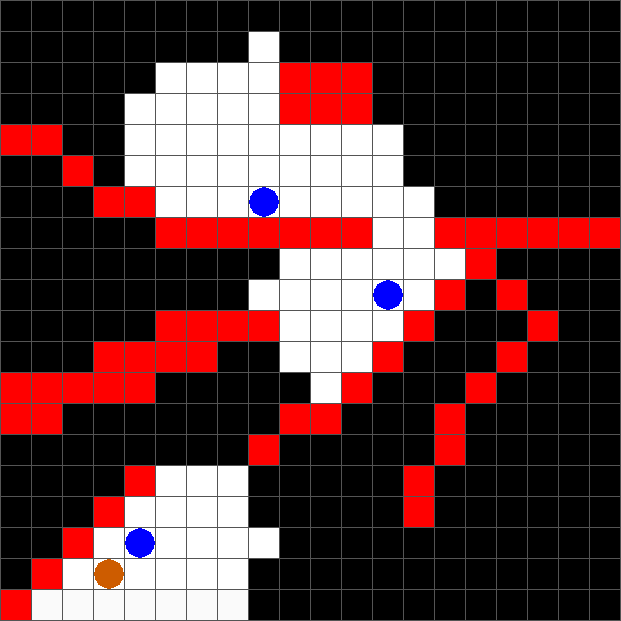}\hspace{.7cm}
		}
		
	\end{minipage}

	\caption{Figures \ref{fig:case1t2} - \ref{fig:case1t7} are chronological snapshots during a simulation of the surveillance game in Figure \ref{fig:3experiment}. Grey regions represent the global belief of the target's location.  
	}\vspace{-0.5cm}
	\label{fig:case1exp}
	
\end{figure}

\begin{table}[h!]
\vspace{-0.2cm}
	\centering
	\caption{Synthesis times for each surveillance subgame}
	\label{tab:synthtime}
	\begin{tabular}{c|l|l|l}
		\multicolumn{1}{l|}{}                                    & \textbf{Subgame} & \textbf{Number of states} & \textbf{Synthesis time (s)} \\ \hline \hline
		\multirow{7}{*}{\textbf{6 sensors}}
		& Subgame 1   & 69     & 101                          \\
	    & Subgame 2   & 74     & 206                          \\
		& Subgame 3   & 62     & 111                          \\
		& Subgame 4   & 52     & 88                          \\
		& Subgame 5   & 77     & 285                          \\
		& Subgame 6   & 66     & 64                          \\ \hline
		& \textbf{Total}   & \textbf{400}         & \textbf{855}                         \\ \hline
		\multicolumn{1}{l|}{\multirow{4}{*}{\textbf{3 sensors}}} & Subgame 1        & 142 & 473                         \\
		\multicolumn{1}{l|}{}                                    & Subgame 2        & 113 & 306                         \\
		\multicolumn{1}{l|}{}                                    & Subgame 3        & 145 & 372                         \\ \hline
		\multicolumn{1}{l|}{}                                    &  \textbf{Total} & \textbf{400}            & \textbf{1151}                        
	\end{tabular}
\end{table}
\vspace{-0.2cm}
The multi-agent surveillance game in Figure \ref{fig:experiment} results in more subgames compared to the game in \ref{fig:3experiment}. However, each game is much smaller and strategies can be synthesized faster in each subgame. Figure \ref{fig:case1exp} shows snapshots in time of the simulation of the 3 sensor surveillance game in Figure \ref{fig:3experiment}. The target is being controlled by a human and the sensors are following their synthesized local surveillance strategies. The global belief is depicted in Figure \ref{fig:case1exp} as grey cells, meaning that the combined knowledge of all the sensors has restricted the location of the target into one of the grey cells.

 We see, in Figures \ref{fig:case1t2} - \ref{fig:case1t4}, that the target is in the subgame corresponding to sensor 2. Hence, only sensor 2 is moving and trying to lower its belief to below 5 cells (which it does in Figure \ref{fig:case1t5}). In Figures \ref{fig:case1t3} - \ref{fig:case1t5}, the target starts moving towards subgame 3 at which point the target is detected by the static sensor in subgame 3 and sensor 3 takes over in figures \ref{fig:case1t6} - \ref{fig:case1t7}. There is no coordination between any of the agents, and each satisfy only their local surveillance specification. However, our construction guarantees that the global specification of $\LTLfinally \LTLglobally p_5$ will be satisfied. 
 


\section{CONCLUSIONS}
We presented a method for decentralized synthesis of surveillance strategies for a mobile sensor network working together with static sensors. Problems that would otherwise be computationally intractable can be solved by decomposing the global game into local subgames for each sensor with individual surveillance specifications. We show that although each game is solved completely independently with no information sharing, we can still guarantee global surveillance properties.
In future work we aim to incorporate false positives in static alarm triggers as well as noisy observations from the mobile sensors while still guaranteeing surveillance specifications.

{\bf Acknowledgement:} This work was supported in part by grant Sandia National Lab 801KOB, grant ARO W911NF-15-1-0592,  and grant DARPA W911NF-16-1-0001.





\bibliographystyle{IEEEtran}
\bibliography{main}

\end{document}